# RIFT: Multi-modal Image Matching Based on Radiation-invariant Feature Transform

Jiayuan Li, Qingwu Hu, and Mingyao Ai

*Abstract*—Traditional feature matching methods such as scale-invariant feature transform (SIFT) usually use image intensity or gradient information to detect and describe feature points; however, both intensity and gradient are sensitive to nonlinear radiation distortions (NRD). To solve the problem, this paper proposes a novel feature matching algorithm that is robust to large NRD. The proposed method is called radiation-invariant feature transform (RIFT). There are three main contributions in RIFT: first, RIFT uses phase congruency (PC) instead of image intensity for feature point detection. RIFT considers both the number and repeatability of feature points, and detects both corner points and edge points on the PC map. Second, RIFT originally proposes a maximum index map (MIM) for feature description. MIM is constructed from the log-Gabor convolution sequence and is much more robust to NRD than traditional gradient map. Thus, RIFT not only largely improves the stability of feature detection, but also overcomes the limitation of gradient information for feature description. Third, RIFT analyzes the inherent influence of rotations on the values of MIM, and realizes rotation invariance. We use six different types of multi-model image datasets to evaluate RIFT, including optical-optical, infrared-optical, synthetic aperture radar (SAR)-optical, depth-optical, map-optical, and day-night datasets. Experimental results show that RIFT is much more superior to SIFT and SAR-SIFT. To the best of our knowledge, RIFT is the first feature matching algorithm that can achieve good performance on all the above-mentioned types of multi-model images. The source code of RIFT and multi-modal remote sensing image datasets are made public[1].

## I. INTRODUCTION

IMAGE feature matching is a fundamental and crucial issue in photogrammetry and remote sensing, whose goal is to extract reliable feature correspondences from two or more images with overlapping regions. It is also widely used in the fields of computer vision, artificial intelligence, robot vision, medical image analysis and so on. Image matching has always been a hot research issue and made great progress in the past decades. However, image matching, especially remote sensing image matching, is still an ill-posed problem, which suffers from many uncertainties. In general, a remote sensing image pair may contain scale, rotation, radiance, noise, blur, or temporal changes. These huge differences in geometry and radiation pose a daunting challenge to the current image matching algorithms, which result in a substantial reduction in matching performance and make it difficult to meet the requirements of ever-changing practical applications. Therefore, it is very important to study more effective, universal, and robust image matching algorithms. To achieve universal robust image matching, three difficult problems need to be solved: (1) the robustness to various geometric and radiation distortions; (2) more efficient and robust outlier detection models; (3) non-rigid deformation image matching problem. In our past work, we have studied geometric distortions [1], non-rigid matching [2], and outlier removal [3-5]; in this paper, we will focus on radiation distortions, especially nonlinear radiation distortions (NRD).

Radiation distortions refer to the phenomenon that the spectral emissivity of the ground objects is different from the real spectral emissivity in the process of sensor imaging. The factors that cause radiation distortions are various, which can be mainly summarized as two aspects: (1) the imaging property of the sensor itself. This kind of error can be regarded as systematic error. Images acquired by the same sensor often have the same systematic error, and thus have little effect on the classical image matching algorithms. However, with the diversification of sensors and applications, it is often necessary to fuse the superiority information of different sensors, and finally achieve a more accurate and reliable description of the scene. The imaging mechanism of different sensors may be quite different, and the acquired images have different expressions for the same objects, which result in large radiation differences between image pairs. For linear radiation differences, classical feature matching methods are still able to deal with; however, for NRD, these methods may not work. Generally, the images with large NRD are called multi-modal images. Traditional methods usually use intensity information or gradient information for feature detection and description. However, both image intensity and gradient are very sensitive to NRD. It can be said that the processing of multi-modal images is a bottleneck problem of image matching. At present, if the geometrical geographic information of images is unavailable, no image matching method can be simultaneously applied to optical-optical matching, infrared-optical matching, synthetic aperture radar (SAR)-optical matching, depth-optical matching, map-optical matching, and day-night matching. (2) the radiation transmission error caused by the atmosphere. In the process of electromagnetic wave transmission, the spectral emissivity of ground objects may be distorted by the influence of atmospheric action, solar altitude angle, and illumination conditions. This kind of distortions is especially prominent in multi-temporal remote sensing images, which often appears "different objects with same spectrum" or "same object with different spectra" phenomenon. Such nonlinear differences will reduce the correlation between correspondences, which often results in difficulties in matching. Because multi-modal and multi-temporal remote sensing image data play an important role in target detection, disaster assessment, illegal building detection, and land resource change monitoring, it is imperative to study image matching methods against large radiation differences, especially large NRD.

In this paper, we propose a radiation-invariant feature matching method based on phase congruency (PC) and maximum index map (MIM), which is called radiation-invariant feature transform (RIFT). First, we detect corner feature points and edge feature points on the PC map. We find that corner feature usually has better repeatability and edge feature has more number. Thus, the combination of corner features and edge features can improve stability of feature detection. Then, we construct a MIM based on log-Gabor convolution sequence, which is much more robust to NRD than traditional gradient map. Thus, MIM is very suitable for multi-modal image feature description. However, MIM is very sensitive to rotations. Different rotation angles may result in different MIMs. We analyze the inherent influence of rotations on the values of MIM, and achieve rotation-invariant by construction of multiple MIMs. Experiments show that RIFT is far superior to scale invariant feature transform (SIFT) [6] and SAR-SIFT [7]. To the best of our knowledge, RIFT is the first feature matching algorithm that can achieve good performance on all different types of multi-model images.

---
[1] https://sites.google.com/site/jiayuanli2016whu/home

## II. Related work

Image matching is one of the most important steps in the automatic production of photogrammetry and remote sensing products, and its results directly affect the applications of image stitching, bundle adjustment, and 3D reconstruction. The literature [8] makes a very systematic summary and classification of traditional image matching methods. According to it, image matching methods can be divided into two categories: area-based matching methods and feature-based matching methods.

### A. Area-based Methods

Area-based matching methods use the original pixel values and specific similarity measures to find the matching relationship between image pairs. Usually, a predefined local window or global image is used to search for matching, and no feature extraction stage is required [9].

One of the drawbacks of area-based methods is that they are only suitable to image pairs containing translation changes. For image pairs containing rotation changes, a circular window is needed to perform the correlation. However, if the image pairs contain complex variations such as rotations, scale changes, geometric distortions, etc., these methods will fail.

Another drawback is that the content inside the local window is not salient. The image content inside the search window may be relatively smooth, and lack of salient features. If the search window is located in a weakly textured or non-textured area of the image, then this method is likely to get an incorrect match. Therefore, the selection of window should be based on the image content, and the portion containing more salient features should be selected as the search window content.

Area-based methods can be roughly divided into three sub-categories, including correlation-based methods [10, 11], Fourier-based methods [12, 13], and mutual information-based methods [14, 15].

### B. Feature-based Methods

Different from area-based methods, feature-based methods are not directly based on image intensity information. These methods usually first detect salient structural features in the image, which can be point features (corner points, straight line intersections, etc.), line features (straight line, contour line, etc.), or region features (building, lake, etc.). However, these features must be salient, easily detectable, and stable. That is to say, no matter how effect of image geometric distortions and radiation distortions, there are always enough identical elements in the two feature sets. In the following, we will only review point feature matching methods, because point feature is the simplest and the base of other features.

Features can better describe the structure information of an image, thus reducing the sensitivity to the sensor noise and scene noise. Feature-based methods are generally more robust than area-based methods. In the field of computer vision, SIFT [6] is one of the most commonly used and effective feature point matching methods. It first constructs a Gaussian scale space and extracts feature points in the scale space. Then, SIFT uses a gradient histogram technique to describe features. Speeded-up robust features (SURF) [16] extracts feature points based on Hessian matrix and introduces integration graph technique to improve efficiency. Principal component analysis SIFT (PCA-SIFT) [17] adopts the principal component analysis algorithm to reduce dimensionality of SIFT descriptor, and extracts the dimensions with larger values. PCA-SIFT greatly reduces the complexity of the original SIFT. Affine-SIFT (ASIFT) [18] extends SIFT to be invariant to affine transformations by simulating two camera axis direction parameters. ORB (Oriented FAST and rotated BRIEF) [19] algorithm uses features from accelerated segment test (FAST) [20] to extract feature points and utilizes directional binary robust independent elementary features (BRIEF) [21] algorithm for feature description. This method has low time complexity and is suitable for real-time applications, but it is not scale-invariant. In the field of photogrammetry and remote sensing, SIFT algorithm has also been widely adopted due to its robustness to illumination, rotation, scale, noise, and viewpoint changes. However, because remote sensing images are captured by different sensors, at different time, and with different viewpoints, there are large geometric distortions and radiation distortions between remote sensing image pairs. To solve such a problem, scholars have proposed many improved SIFT algorithms. Uniform robust SIFT (UR-SIFT) [22] studies the distribution of SIFT feature points and proposes an entropy-based feature point selection strategy to improve the distribution uniformity. SAR-SIFT [7] introduces a new gradient definition based on the specific characteristics of SAR images to improve the robustness to speckle noise. Adaptive binning SIFT (AB-SIFT) [23] adopts an adaptive binning gradient histogram to describe feature points, making it more resistant to local geometric distortions. However, these feature-based methods usually use gradient information to detect and describe feature points, and are very sensitive to NRD.

This paper aims to solve the problem of radiation distortions in image matching, especially the problem of NRD. Multi-modal images are typical images with NRD. At present, the research of multi-modal image matching mostly focuses on medical images, and the processing of multimode remote sensing images is very few. However, multi-modal remote sensing image matching has very important theoretical and practical significance. Theoretically, this problem is very difficult, and it is a bottleneck problem of image matching technology. In fact, many applications require automatic matching of multi-modal images, such as information fusion of optical and SAR images. In next section, we will briefly review state-of-the-art of multi-modal image matching.

*C. Multi-modal Image Matching*

Currently, the most effective multi-modal image matching algorithm belongs to the histogram of orientated phase congruency (HOPC) [24, 25]. HOPC is originally published at the 2016 ISPRS conference and wins the best paper award [25], and then published in the top journal of remote sensing [24]. All these illustrate the importance and difficulty of multi-modal image matching problem.

HOPC extends the PC model to include not only numerical information but also corresponding orientation information. Then, a modified similarity measure HOPCncc is presented based on the improved PC measure and normalized cross correlation (NCC) [26]. The detailed steps of HOPC are as follows:

(1) Preparing the geometric geographic information corresponding to the input images; and using the relatively accurate geographic information to perform geometric correction on the reference image and the target image, respectively, which is able to initially register the image pair so as to eliminate the obvious rotation and translation differences between the image pair. Then, using ground sample distance (GSD) to resample the images to have the same GSD, which eliminates the scale difference between the image pair.

(2) Using Harris [27] detector to extract feature points from the reference image. To avoid the uneven distribution of feature points, a blocking strategy is adopted.

(3) Once the feature points are obtained, the HOPCncc metric is used to search correspondences for these features on the target image. Specifically, each feature point corresponds to a 20×20 pixel search window on the target image. Based on the idea of template matching, the point in the search window that achieves the highest HOPCncc score is regarded as its correspondence.

(4) Using the projection transformation model to eliminate outliers and obtain relatively pure correspondences.

(5) Using iterative least squares method to further purify the correspondences.

From the steps, we can clearly see three major problems:

(1) HOPC needs to know the exact geographic information corresponding to the image to perform geometric correction. In practice, however, the geographical information of an image may not be accurate enough or even unavailable. For example, the geographical information of satellite images may be hundreds of meters away from its actual geographical location. In such cases, HOPC is completely unusable.

(2) Although HOPC performs feature detection on the reference image, it is essentially a template matching method, rather than a feature matching method, and therefore is sensitive to rotation, scale, etc. Template matching methods usually performs two-dimensional search, and becomes one-dimensional search after adding the epipolar constraint. However, HOPC relies on the accurate geographic information, whose search space is small, usually a local window of 20×20 pixels.

(3) HOPC uses Harris detector to detect the feature points. However, Harris is very sensitive to NRD, and it is difficult to be universally suitable for different types of multi-modal images. Especially, when a depth map is used as the reference image, the performance of Harris becomes very poor. Feature detection is the basis of feature matching method, which determines the number of correct matches between two sets of points and point location accuracy. If the number of features is too small, the matching result must be very poor.

In contrast, RIFT does not rely on the geographic information. RIFT has good robustness to NRD, regardless of feature detection or feature description stage, and achieves rotation invariance. The performance of RIFT is obviously superior to HOPC.

III. RADIATION-INVARIANT FEATURE TRANSFORM (RIFT)

In this section, we will detail the proposed RIFT method, including feature detection and feature description stages.

*A. Feature Detection*

Classical feature matching methods generally rely on image intensity or gradient information, which is spatial domain information. In addition to spatial domain information, images can also be described using frequency domain information, such as phase information. The theoretical basis of phase is Fourier transform (FT) theorem [28]. FT can decompose an image into amplitude component and phase component. Oppenheim and Lim [29], for the first time, revealed the importance of phase information for the preservation of image features. Phase information has high invariance to image contrast, illumination, scale, and other changes. Further, Morrone and Owens [30] discovered that certain points in the image can cause a strong response from human visual system, and these points usually have highly consistent local phase information. Therefore, the degree of consistency of the local phase information at different angles is called PC measure.

*1) Log-Gabor Filter*

2D log-Gabor filter (2D-LGF) [31, 32] can generally be obtained by Gaussian spreading of the vertical direction of log-Gabor filter (LGF). Thus, 2D-LGF function is defined as follows,

$$L(\rho,\theta,s,o) = \exp(\frac{-(\rho-\rho_s)^2}{2\sigma_\rho^2})\exp(\frac{-(\theta-\theta_{so})^2}{2\sigma_\theta^2}) \quad (1)$$

where $(\rho,\theta)$ represents the log-polar coordinates; $s$ and $o$ are the scale and orientation of 2D-LGF, respectively; $(\rho_s,\theta_{so})$ is the center frequency of 2D-LGF; $\sigma_\rho$ and $\sigma_\theta$ are the bandwidths in $\rho$ and $\theta$, respectively.

LGF is a frequency domain filter, whose corresponding spatial domain filter can be obtained by inverse Fourier transform. In spatial domain, 2D-LGF can be represented as [24, 33],

$$L(x,y,s,o) = L^{even}(x,y,s,o) + iL^{odd}(x,y,s,o) \quad (2)$$

where the real part $L^{even}(x,y,s,o)$ and the imaginary part $L^{odd}(x,y,s,o)$ stand for the even-symmetric and the odd-symmetric log-Gabor wavelets, respectively.

*2) Phase Congruency (PC)*

Let $I(x, y)$ denotes a 2D image signal. Convolving the image $I(x, y)$ with the even-symmetric and the odd-symmetric wavelets yields the response components $E_{so}(x, y)$ and $O_{so}(x, y)$,

$$[E_{so}(x,y), O_{so}(x,y)] = [I(x,y)*L^{even}(x,y,s,o), I(x,y)*L^{odd}(x,y,s,o)] \quad (3)$$

Then, the amplitude component $A_{so}(x, y)$ and phase component $\phi_{so}(x, y)$ of $I(x, y)$ at scale $s$ and orientation $o$ can be obtained by,

$$A_{so}(x, y) = \sqrt{E_{so}(x, y)^2 + O_{so}(x, y)^2} \quad (4)$$

$$\phi_{so}(x, y) = \arctan(O_{so}(x, y) / E_{so}(x, y)) \quad (5)$$

Considering the analysis results in all directions and all orientations, and introducing the noise compensation term $T$, the final 2D PC model is [34]:

$$PC(x, y) = \frac{\sum_s \sum_o w_o(x, y) \lfloor A_{so}(x, y) \Delta \Phi_{so}(x, y) - T \rfloor}{\sum_s \sum_o A_{so}(x, y) + \xi} \quad (6)$$

where $w_o(x, y)$ is a weighting function; $\xi$ is a small value; $\lfloor \cdot \rfloor$ operator prevents the enclosed quantity from getting a negative value; that is, it takes zero when the enclosed quantity is negative. $\Delta \Phi_{so}(x, y)$ is a phase deviation function, whose definition is,

$$A_{so}(x,y)\Delta\Phi_{so}(x,y) = (E_{so}(x,y)\overline{\phi}_E(x,y) + O_{so}(x,y)\overline{\phi}_O(x,y)) \\ - |(E_{so}(x,y)\overline{\phi}_O(x,y) - O_{so}(x,y)\overline{\phi}_E(x,y))| \quad (7)$$

where,

$$\overline{\phi}_E(x, y) = \sum_s \sum_o E_{so}(x, y) / C(x, y) \quad (8)$$

$$\overline{\phi}_O(x, y) = \sum_s \sum_o O_{so}(x, y) / C(x, y) \quad (9)$$

$$C(x, y) = \sqrt{(\sum_s \sum_o E_{so}(x,y))^2 + (\sum_s \sum_o O_{so}(x,y))^2} \quad (10)$$

*3) Corner and Edge Features*

Based on equation (6), we can obtain a very precise edge map, i.e., the PC map. However, this formula ignores the effect of orientation changes on the PC measure [35]. To get the relations between PC measure and orientation changes, we compute an independent PC map $PC(\theta_o)$ for each orientation $o$. We then calculate moments of these PC maps and analyze moment changes with orientations.

According to the moment analysis algorithm [36], the axis corresponding to the minimum moment is called the principal axis, and the principal axis usually indicates the direction information of the feature; the axis corresponding to the maximum moment is perpendicular to the principal axis, and the magnitude of the maximum moment generally reflects the distinctiveness of the feature. Before calculation of the minimum and maximum moments, we first compute three intermediate quantities,

$$a = \sum_o (PC(\theta_o)\cos(\theta_o))^2 \quad (11)$$

$$b = 2\sum_o (PC(\theta_o)\cos(\theta_o))(PC(\theta_o)\sin(\theta_o)) \quad (12)$$

$$c = \sum_o (PC(\theta_o)\sin(\theta_o))^2 \quad (13)$$

Then, the principal axis $\psi$, minimum moment $M_\psi$, and maximum moment $m_\psi$ are given by,

$$\psi = \frac{1}{2}\arctan(\frac{b}{a-c}) \quad (14)$$

$$M_\psi = \frac{1}{2}(c + a + \sqrt{b^2 + (a-c)^2}) \quad (15)$$

$$m_\psi = \frac{1}{2}(c + a - \sqrt{b^2 + (a-c)^2}) \quad (16)$$

The minimum moment $m_\psi$ is equivalent to the cornerness measure in corner detector. In other words, if the value of $m_\psi$ at a point is large, the point is most likely to be a corner feature; and the maximum moment $M_\psi$ is the edge map of an image, which can be used for edge feature detection. Specifically, we first compute $m_\psi$ and $M_\psi$ of the PC maps. For the minimum moment map $m_\psi$, local maxima detection and non-maximal suppression are performed, and the remaining points are accepted as corner features. Because edge structure information has better resistance to radiation distortions, we also use the maximum moment map $M_\psi$ to detect edge feature points, that is, perform FAST feature detection on $M_\psi$ (Noted that other feature detectors can also be used here. Using FAST only considers its time efficiency). Therefore, the proposed method integrates corner features and edge features for feature matching.

Fig. 1 shows an example of feature detection, where Fig. 1(a) is a pair of multi-modal images (an optical satellite image and a LIDAR depth map); Fig. 1(b) and Fig. 1(c) are the minimum moment map $m_\psi$ and maximum moment map $M_\psi$, respectively; Fig. 1(d) is the result of FAST feature detection; Fig. 1(e) and Fig. 1(f) are our corner features and edge features, respectively. From the results, we can draw several conclusions: (1) Comparing Fig. 1(b) and Fig. 1(f), we can find that traditional feature detectors based on image intensity or gradient (such as FAST or Harris detectors) are very sensitive to NRD, while the moments of PC measures have good invariance to NRD. A large number of reliable feature points can be obtained by performing the same FAST or Harris detector on the maximum moment of PC maps. (2) From Fig. 1(d), it can be seen that the obvious corner features can be obtained, and the feature repeatability is high. However, the number of feature points is relatively small. (3) From Fig. 1(f), it can be seen that more feature points can be detected on the maximum moment map, but the repeatability is relatively low.

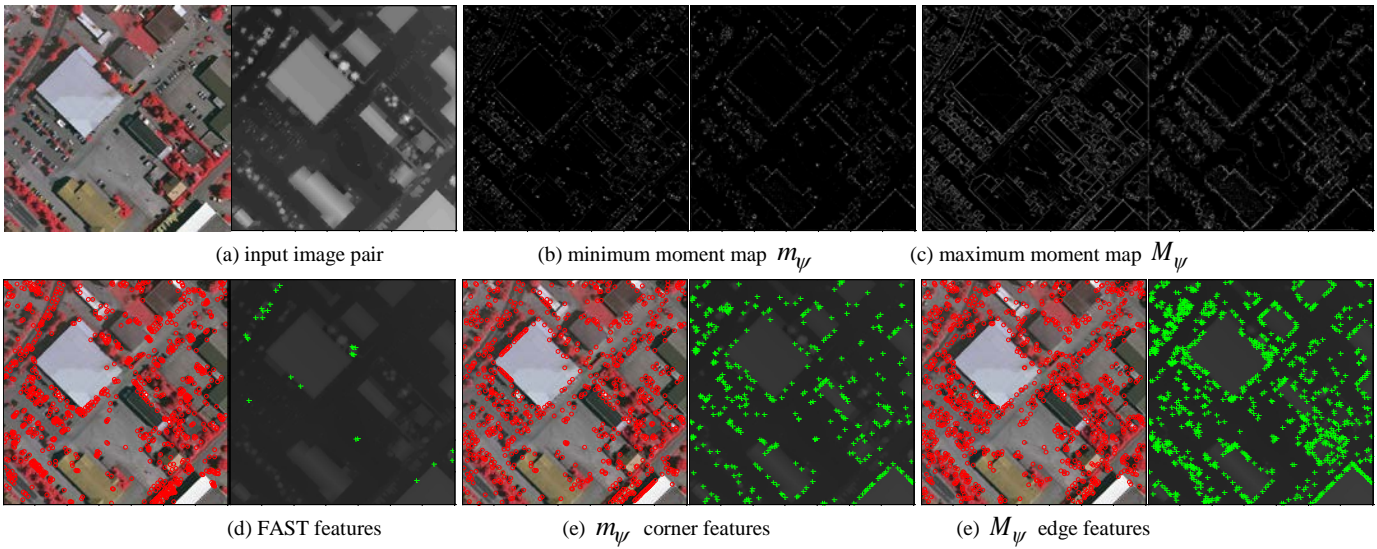

(a) input image pair   (b) minimum moment map $m_\psi$   (c) maximum moment map $M_\psi$

(d) FAST features   (e) $m_\psi$ corner features   (e) $M_\psi$ edge features

Fig. 1. Feature detection.

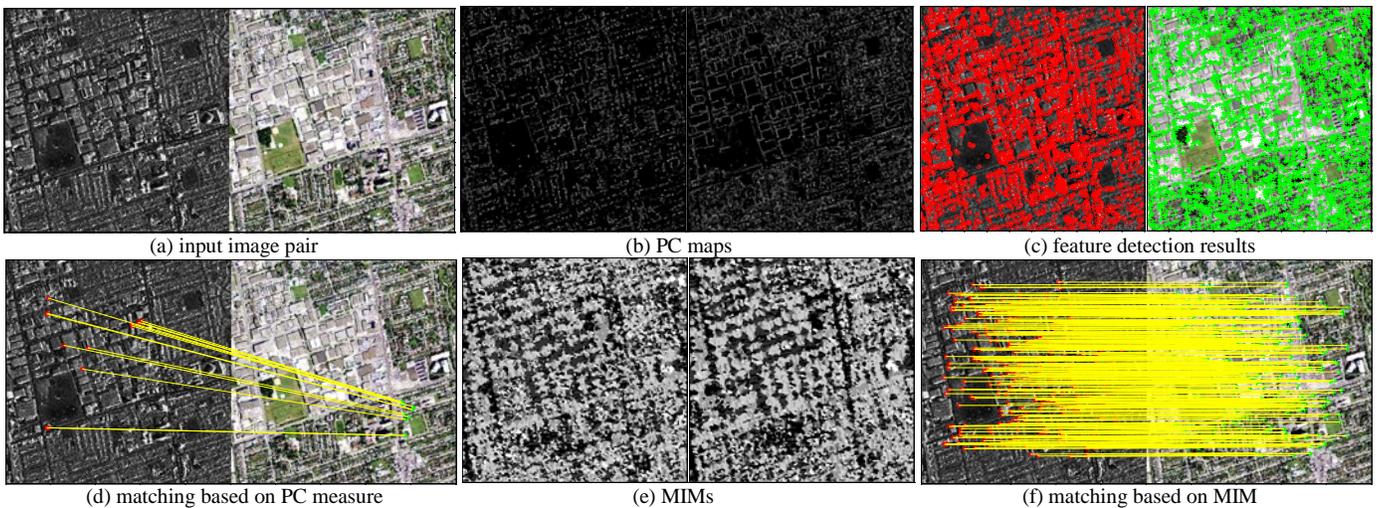

(a) input image pair   (b) PC maps   (c) feature detection results

(d) matching based on PC measure   (e) MIMs   (f) matching based on MIM

Fig. 2. Comparision of MIM with PC.

Therefore, the combination of the characteristics of the minimum moment map corner features and the maximum moment map edge features not only ensures that the high repeatability of the features, but also that the large number of the features, which lays a foundation for subsequent feature matching (both the matching precision and the number of matches).

### B. Feature Description

Once feature points are detected, feature description is needed to increase the distinction between features. Classical feature descriptors generally use image intensity or gradient distribution to construct feature vectors. However, as mentioned earlier, both intensity and gradient are very sensitive to NRD. These descriptors are not suitable for multi-modal image matching task. In the above, we have analyzed the characteristics of PC measure, whose advantage is the robustness to NRD. Intuitively, using PC map instead of intensity map or gradient map for feature description is more suitable. However, experimental results do not reach our expectations. Specifically, we select a remote sensing image pair that is composed of a SAR satellite image and an optical satellite image for test (see in Fig. 2(a)). We first compute the PC maps (see in Fig. 2(b)) and detect corner and edge features from each image (see in Fig. 2(c)). Then, for each feature, we construct a 6×6×6=216 dimensional feature vector based on distribution histogram technique. The matching result based on PC map description is shown in Fig. 2(d).

From Fig. 2, we can know that the number and distribution of extracted features are quite good; however, the matching result is very poor, where the matches are almost all outliers. It shows that the PC map is not quite suitable for feature description. The reasons may be as follows: first, the information of PC map is less, since most pixel values in PC map are close to zero. It is not robust enough for feature description. Second, PC map is sensitive to noise because it mainly contains edges, which causes the feature description to be inaccurate. With such analyses, we present a MIM measure instead of PC map for feature description.

#### 1) Maximum Index Map (MIM)

MIM is constructed via log-Gabor convolution sequence. The convolution sequence has already been obtained in the PC map calculation stage. Therefore, the computation complexity of MIM is very small. Fig. 3 illustrates the construction of MIM.

Given an image $I(x, y)$, we first convolve $I(x, y)$ with a 2D-LGF to obtain the response components $E_{so}(x, y)$ and $O_{so}(x, y)$; and then calculate the amplitude $A_{so}(x, y)$ at scale $s$ and orientation $o$. For each orientation $o$, the amplitudes of all $N_s$ scales are summed up to get the log-Gabor convolution layer $A_o(x, y)$,

$$A_o(x, y) = \sum_{n=1}^{N_s} A_{so}(x, y) \qquad (17)$$

The log-Gabor convolution sequence is obtained by arranging the log-Gabor convolution layers in order, which is a multi-channel convolution map $\{A_o^\omega(x, y)\}_1^{N_o}$, where $N_o$ is the number of orientations; the superscript $\omega = 1, 2, ..., N_o$ represents different channels of the log-Gabor convolution sequence. Thus, for each pixel position $(x_j, y_j)$ of the convolution map, we can get a $N_o$-dimensional ordered array $\{A_o^\omega(x_j, y_j)\}_1^{N_o}$, and then find the maximum value $A_{\max}(x_j, y_j)$ in this array. The channel where $A_{\max}(x_j, y_j)$ locates is denoted by $\omega_{\max}$, which is the maximum index value of this pixel position. We set $\omega_{\max}$ as the pixel value of position $(x_j, y_j)$ in MIM.

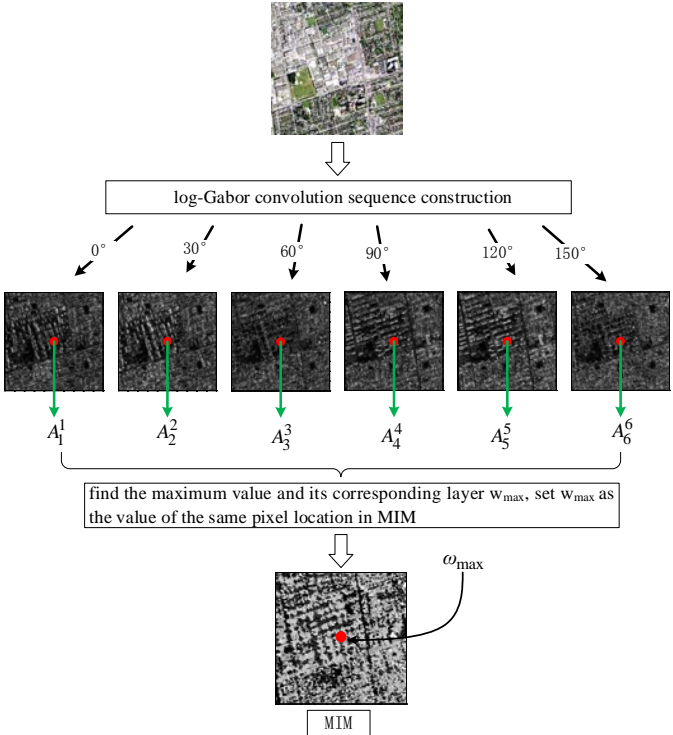

Fig. 3. The construction of MIM.

After obtaining the MIM, we use a distribution histogram technique similar to SIFT for feature vector description. In detail, for each feature point, we select a local image patch with $J \times J$ pixels centered at the feature, and use a Gaussian function whose standard deviation is equal to $J/2$ to assign weights for each pixel. This process is to avoid abrupt changes in feature description if the window position changes. We then divide the local patch into 6x6 sub-grids, and build a distribution histogram with $N_o$ bins for each sub-grids, because the values of MIM range from 1 to $N_o$. The feature vector is obtained by concatenating all the histograms. Thus, the dimension of the feature vector is $6 \times 6 \times N_o$. To gain invariance to illumination changes, we finally normalize the feature vector.

A matching example based on MIM description is given in Fig. 2, where Fig. 2(e) is the MIMs corresponding to Fig. 2(a), and Fig. 2(f) is the matching result. In this experiment, we set $N_o = 6$ and use MIM instead of traditional gradient for feature description. We regard the feature point pairs with minimal Euclidean distance as potential matches and apply NBCS [3] method for outlier removal. As can be seen, the proposed method can extract a large number of reliable matches with relatively uniform distribution even in the SAR and optical image pair. The imaging mechanisms of SAR and optical sensors are quite different, which results in large NRD between SAR image and optical image. Thus, it shows that the proposed MIM descriptor is very suitable for multi-modal image matching task, and is much better than traditional feature matching methods.

*2) Rotation Invariance*

The previous section has analyzed the possibility and validity of the MIM for feature description, and described the feature vector construction details. However, the description method assumes that there are no rotations between an image pair; that is, the rotation changes are not considered. Then, if there is a rotation change in the image pair, the above method will no longer be suitable. Therefore, special processing must be performed to make it rotationally invariant. The most straightforward idea is to use the dominant orientation method similar to SIFT. However, after extensive experiments, we found that rotation invariance can not be achieved by dominant orientation method.

To analyze the reasons, two experiments are performed. Fig. 4 analyzes the affection of rotations on gradient map, where Fig. 4(a) is a LIDAR point cloud depth map; Fig. 4(d) is obtained by rotating Fig. 4(a) clockwise 30°; Fig. 4(b) and Fig. 4(e) are the gradient maps of Fig. 4(a) and Fig. 4(d), respectively. To eliminate the rotation difference between Fig. 4(b) and Fig. 4(e), Fig. 4(b) is rotated clockwise by 30° to obtain Fig. 4(c); Fig. 4(f) is the difference between Fig. 4(b) and Fig. 4(e). According to Fig. 4(f), the gradient maps after removing the rotation difference are basically the same, indicating that the rotation has no influence on values of the gradient map. Therefore, by calculating the main orientation of the feature point, the rotation difference between local image patch can be eliminated, thereby achieving rotation invariance. Similarly, the above analysis is also performed on MIM, as shown in Figure 5. Fig. 5(b) and Fig. 5(e) are the MIMs of Fig. 5(a) and Fig. 5(d), respectively; Fig. 5(c) is obtained by rotating Fig. 5(b) clockwise 30°; Fig. 5(f) is the difference between Fig. 5(b) and Fig. 5(e). Only if Fig. 5(c) is similar enough to Fig. 5(e), the dominant orientation method can be applied. However, most of the values of Fig. 5(f) are not close to zero, indicating that there is not only a rotation difference between Fig. 5(c) and Fig. 5(e), but also a numerical difference, and this numerical difference is caused by rotations. Thus, to achieve rotation invariance, we must figure out the relationship between rotations and the values of MIM.

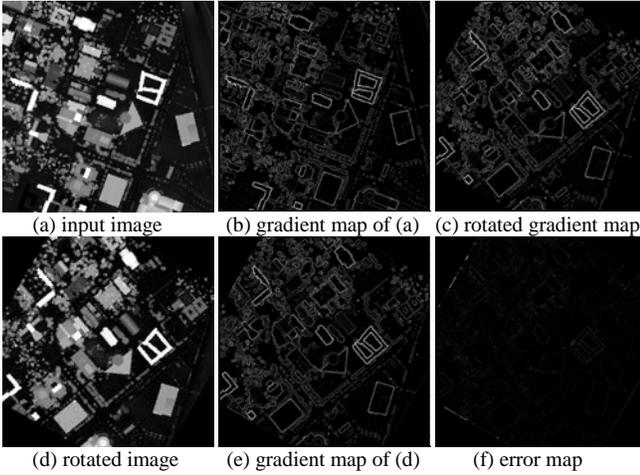

(a) input image    (b) gradient map of (a)    (c) rotated gradient map
(d) rotated image    (e) gradient map of (d)    (f) error map
Fig. 4. The affection of rotations on gradient map.

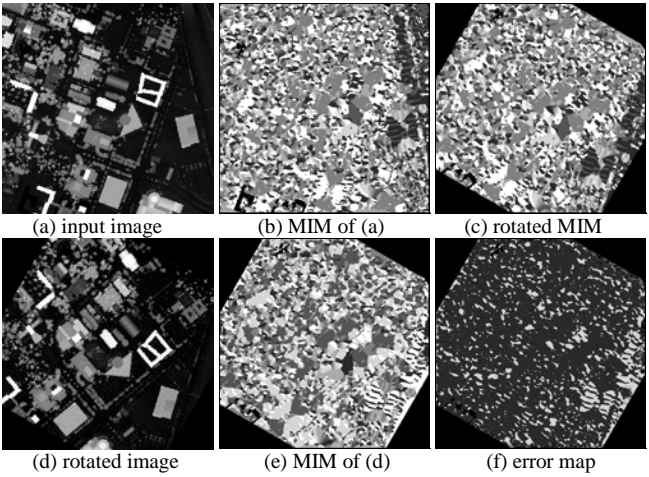

(a) input image    (b) MIM of (a)    (c) rotated MIM
(d) rotated image    (e) MIM of (d)    (f) error map
Fig. 5. The affection of rotations on MIM.

As abovementioned, MIM is constructed based on log-Gabor convolution sequence, and the convolution layer is closely related to orientations. Therefore, if the start layer of the log-Gabor convolution sequence is different, then the constructed MIM is completely different.

In other words, if two images are to be successfully matched, the log-Gabor convolution sequences corresponding to the two images must have high similarities, and each layer of the log-Gabor convolution sequence needs to be similar. In fact, the log-Gabor convolution sequence can be thought of as an end-to-end annular structure, as shown in Fig. 6. Assume that Fig. 6(a) is a 6-layers log-Gabor convolution sequence ring (noted by $S_A$) obtained from the original image (image in Fig. 4(a)), where the first layer is the 0° direction convolution result (the initial layer of the convolution sequence); the second layer is the 30° direction convolution result, and so on. The sixth layer is the 150° direction convolution result. However, if we rotate the image by an angle (as shown in Fig. 6(b)), and still use the 0° direction convolution result as the initial layer to construct the convolution sequence (obtaining convolution sequence $S_B$). Due to the effect of rotation, the content of the initial layer of $S_A$ will be quite different from $S_B$. In fact, which layer should be used as the initial layer is not known, because it is highly related to the rotation angle. Considering that $N_o$ is small and generally set to 6, we use the simplest traversal strategy, listing all possible scenarios. In detail, we first construct a convolution sequence $S_A$ and a convolution sequence $S_B$ for the reference image and the target image, respectively. For $S_A$ of the reference image, we directly construct a MIM ($MIM^{S_A}$); for $S_B$ of the target image, we successively transform the initial layer of $S_B$ to reconstruct a set of convolution sequences $\{S_w^B\}_1^{N_o}$ with different initial layers, and then calculate a MIM from each convolution sequence to get a set of MIMs $\{MIM_w^{S_B}\}_1^{N_o}$. In general, there is always a MIM in set $\{MIM_w^{S_B}\}_1^{N_o}$ that is similar to $MIM^{S_A}$. To verify this conclusion more intuitively, we perform an experiment on Fig. 4(d) to obtain MIM set $\{MIM_w^{S_B}\}_1^{N_o}$ ($N_o = 6$). Fig. 7 shows all the MIMs in the set $\{MIM_w^{S_B}\}_1^{N_o}$. The initial layers of $\{S_w^B\}_1^{N_o}$ are the first to sixth layers of the convolution sequence $S_B$. As can be seen, if the initial layers are different, the resulting MIM is completely different. Fig. 8 shows the difference between each subfigure in Fig. 7 and Fig. 5(c). It is found that when the sixth layer is used as the initial layer, the constructed MIM $MIM_6^{S_B}$ is very consistent with $MIM^{S_A}$, which verifies the above conclusion.

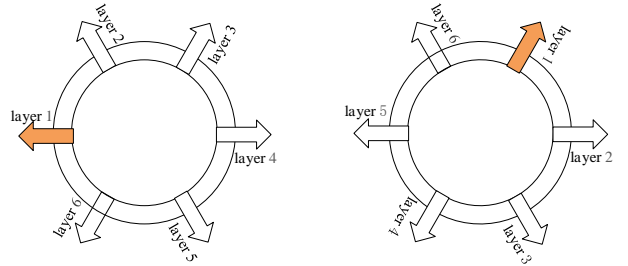

(a) convolution sequence $S_A$    (b) convolution sequence $S_B$
Fig. 6. Convolution sequence ring.

The above process substantially eliminates the effect of rotations on the values of MIM. Then, the dominant orientation method can be directly applied to gain rotation invariance. In summary, the proposed RIFT algorithm builds a feature vector for each keypoint of the reference image and $N_o$ feature vectors for each keypoint of the target image.

## IV. EXPERIMENTAL RESULTS

To verify the effectiveness of the proposed RIFT method, we select several multi-modal data for qualitative and quantitative evaluation. We compare our RIFT algorithm against two state-of-the-arts, i.e., SIFT algorithm and SAR-SIFT algorithm. The parameters of SIFT and SAR-SIFT are set according to their original literature and are consistent in all experiments. For fair comparison, both the implementations of SIFT and SAR-SIFT are obtained from their authors' personal website.

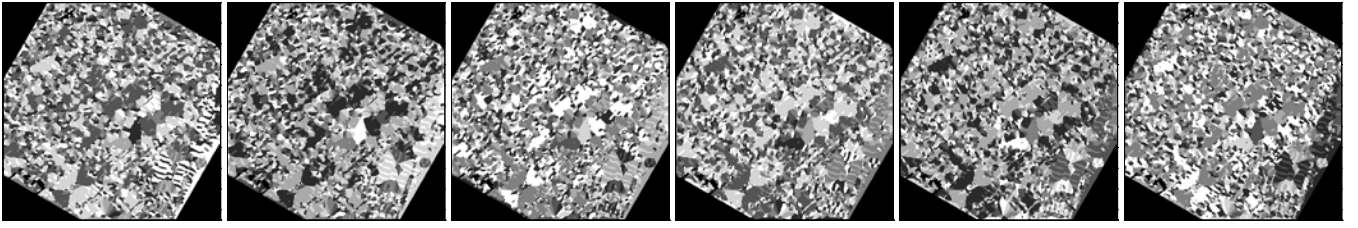

Fig. 7. MIM set $\{MIM_w^{S_B}\}_1^{N_o}$ ( $N_O = 6$ ). The initial layers of are the first to sixth layers of the convolution sequence $S_B$ .

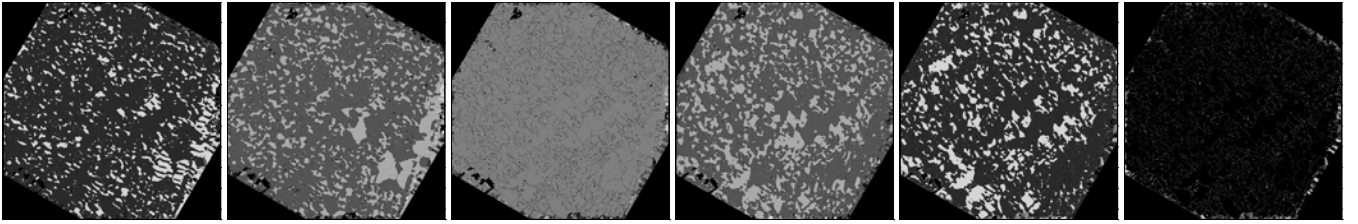

Fig. 8. Error map between each subfigure in Fig. 7 and Fig. 5(c).

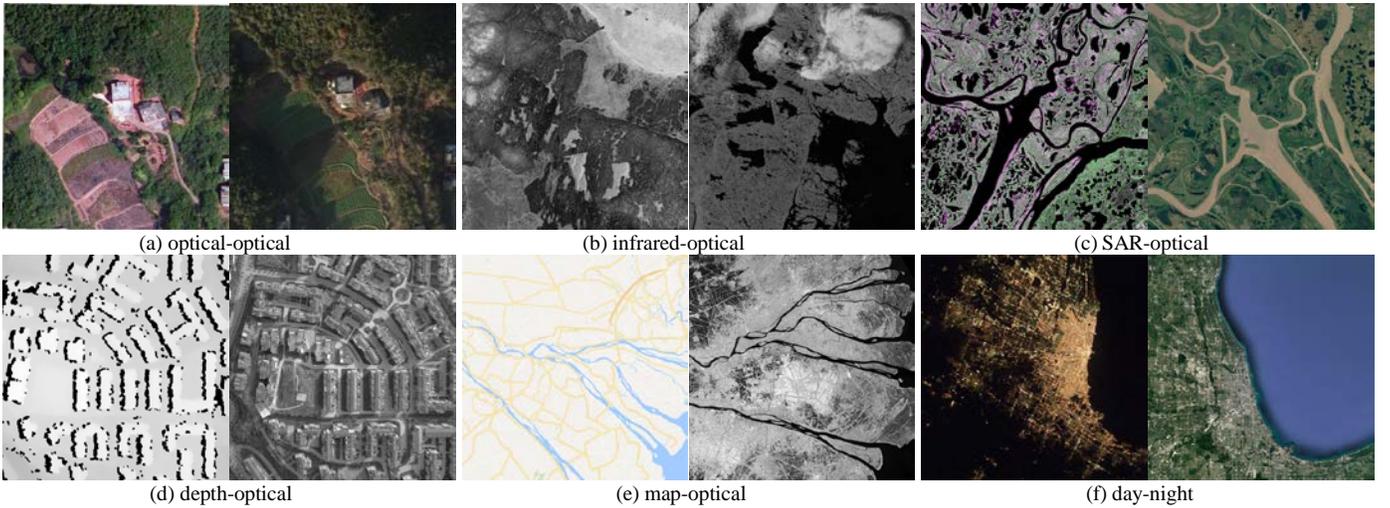

(a) optical-optical　　　　　　　(b) infrared-optical　　　　　　　(c) SAR-optical

(d) depth-optical　　　　　　　(e) map-optical　　　　　　　(f) day-night

Fig. 9. Sample data.

## A. Datasets

Six types of multi-modal image data sets are selected as experimental sets, including optical-optical, infrared-optical, SAR-optical, depth-optical, map-optical, and day-night. Each type of datasets contains 10 image pairs for a total of 60 multi-modal image pairs. The sample data are shown in Fig. 9.

These image pairs include not only multi-sensor images, multi-temporal images, but even artificially produced images, such as rasterized map data; not only images under good lighting conditions (daytime images), but also nighttime remote sensing images; not only high spatial resolution images, but also low and medium spatial resolution images, whose GSD ranges from 0.1 meters to hundreds of meters; not only satellite images, but also UAV images and even close-range images; not only urban area images, but also countryside and even mountain forest images. There are serious distortions between these image pairs, especially radiation distortions, which will bring great challenges into image matching algorithms. Such challenges can test the validity and robustness of the proposed RIFT algorithm more comprehensively. It should be noted that RIFT is currently not scale invariant. Therefore, the two images of each image pair need to be resampled to have approximately the same GSD.

For better quantitative evaluation, we need to obtain a ground truth geometric transformation between each image pair. However, due to the interference of various factors, the real datasets usually do not have a true ground truth geometric transformation. The approximate ground truth geometric transformation is generally used for evaluation. Specifically, for each image pair, we select five evenly distributed correspondences with sub-pixel accuracy and use these correspondences to estimate an accurate affine transformation as the approximation of ground truth geometric transformation. We first perform feature matching on this image pair (RIFT/SIFT/SAR-SIFT), and remove outliers based on NBCS [3] method; then, calculate the residuals of these image correspondences under the estimated affine transform, and regard the correspondences with residuals less than 3 pixels as the correct matches. We use number of correct matches (NCM), root mean square error (RMSE), mean error (ME), and success rate (SR) as the evaluation metrics. Note that if the NCM of an image pair is less than four, the matching is considered to have failed.

## B. Parameter Study

Table 1 The details of parameter settings.

| experiments | variable | fixed parameters |
|---|---|---|
| parameter $N_o$ | $N_o = [4, 5, 6, 7, 8]$ | $N_s = 3$, $J = 96$ |
| parameter $N_s$ | $N_s = [2, 3, 4, 5, 6]$ | $N_o = 6$, $J = 96$ |
| parameter $J$ | $J = [48, 72, 96, 120, 144]$ | $N_o = 6$, $N_s = 3$ |

The proposed RIFT method contains three main parameters, namely, $N_s$, $N_o$, and $J$. Parameter is $N_s$ the number of convolution scales of the log-Gabor filter, and its value is usually greater than 1. Parameter $N_o$ is the number of convolution orientations of the log-Gabor filter. In general, the more the number of orientations is, the richer the amount of information of the constructed MIM gets, and the higher the computational complexity is. Parameter $J$ is the size of the local image patch used for feature description. If the local patch is too small, it contains too less information, which is difficult to reflect the distinctiveness of the feature. On the contrary, if the image patch is too large, it is easily affected by the local geometric distortion. Therefore, suitable parameters are very important. This section performs parameter study and sensitivity analysis based on map-optical dataset. We design three independent experiments to learn parameters $N_s$, $N_o$, and $J$, where each experiment has only one parameter as a variable, and other parameters are fixed values. The experimental setup details are summarized in Table 1. For each parameter, we use NCM and SR as the evaluation metrics. The experimental results are reported in Table 2~Table 4.

From the experimental results, we can infer that: (1) larger values of $N_o$ mean that richer information of the constructed MIM, and thus more NCM can be obtained; however, larger values of $N_o$ also mean that the number of convolution sequences increases, which will greatly increase the computational complexity of the algorithm. From Table 2, when $N_o$ reaches 6, the SR of RIFT reaches 100%. However, increasing the number of orientations only slightly improves the NCM. Therefore, to take into account both the matching performance and computational complexity of RIFT, we set $N_o$ to 6. (2) from Table 3, we can learn that small values of $N_s$ result in low SR accuracy and large values of $N_s$ result in poor NCM performance. When $N_s = 4$, RIFT achieves the best in both SR and NCM metrics. Although the results of $N_s = 3$ are only slightly different from the results of $N_s = 4$, the number of scales is different from the number of directions, increasing the scales do not significantly increase the computational complexity. Therefore, we set $N_s$ to 4. (3) the influence of the parameter $J$ on RIFT is similar to $N_s$. If the value of $J$ is small, the amount of information is not rich enough, the SR and NCM metrics will be poor; however, if the value of $J$ is large, due to the effect of local geometric distortions, the performance of NCM will decrease. As shown in Table 4, RIFT achieves the best performance when $J = 96$. Based on the experimental results and analyses, these parameters are fixed to $N_o = 6$, $N_s = 4$, $J = 96$ in the following experiments.

Table 2 The results of parameter $N_o$.

| metric | $N_o$, $N_s = 3$, $J = 96$ | | | | |
|---|---|---|---|---|---|
| | 4 | 5 | 6 | 7 | 8 |
| NCM | 50.4 | 84.9 | 114.8 | 120.5 | 121.4 |
| SR/% | 60 | 70 | 100 | 100 | 100 |

Table 3 The results of parameter $N_s$.

| metric | $N_s$, $N_o = 6$, $J = 96$ | | | | |
|---|---|---|---|---|---|
| | 2 | 3 | 4 | 5 | 6 |
| NCM | 81 | 114.8 | 119.8 | 102.5 | 89.6 |
| SR/% | 80 | 100 | 100 | 100 | 100 |

Table 4 The results of parameter $J$.

| metric | $J$, $N_o = 6$, $N_s = 3$ | | | | |
|---|---|---|---|---|---|
| | 48 | 72 | 96 | 120 | 144 |
| NCM | 91.9 | 111.6 | 119.8 | 116.6 | 98.7 |
| SR/% | 60 | 100 | 100 | 100 | 100 |

## C. Rotation Invariance Test

Rotation invariance is an important property of the proposed RIFT, which is also a major advantage compared with HOPC method. The calculation of MIMs and PC maps are both related to orientations. The proposed RIFT algorithm generally performs log-Gabor convolution filtering along six directions, i.e., 0°, 30°, 60°, 90°, 120°, and 150°. The angles of these directions only range from 0° to 150°, which inevitably raise a doubt: "If the rotation angle between the image pairs is not within this range, is the proposed RIFT still robust?"

In fact, the proposed RIFT has very good rotation invariance, not only for the rotations between [0°~150°], but also for the rotations in the entire 360° range. To verify this conclusion, an image pair was selected from the map-optical dataset for experimentation. This image pair does not suffer from rotation changes. First, we rotate the map of this image pair. The rotation angles are from 0° to 359° with an interval of 5°. Thus, a total of 72 maps are obtained (the rotation angles are $[0°, 5°, 10°, ..., 345°, 350°, 355°, 359°]$ respectively). These 72 maps and the optical image constitute 72 image pairs. Then, these images are processed one by one using RIFT, and their corresponding NCMs are plotted in Fig.10. The red dots in the figure represent the NCMs. It can be clearly seen that although the NCMs under different rotation angles are different, all the NCMs are greater than 40, indicating that the proposed RIFT can successfully match all the image pairs, and the matching SR accuracy is 100%, which also verifies that the proposed RIFT has good rotation invariance for rotations in the entire 360° interval. Meanwhile, the differences in the NCMs also indicate that the dominant orientation calculation of RIFT may not be optimal, and a more robust feature main orientation calculation method will further improve the matching performance of the proposed RIFT, which will become one of our key research topics in the future. Fig. 11 shows the experimental results for 150° rotation and 210° rotation. Among them, the first raw is the results of feature matching (yellow lines in the figure represent correct matches), and the second row is the registration results. It can be seen that the NCMs are large; the distribution of matching points is relatively uniform; and the registration accuracy is very high.

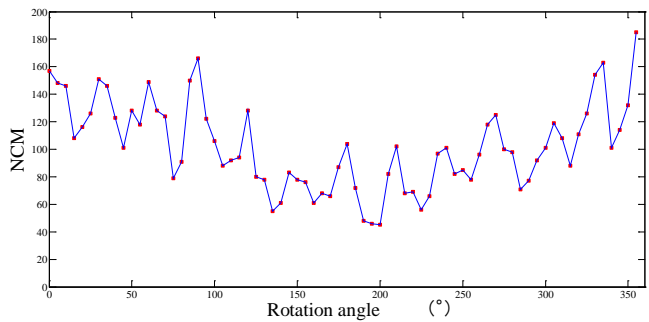

Fig. 10. Rotation invariance test.

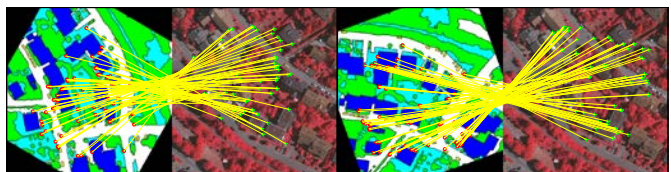

(a) matching result of 150° rotation    (b) matching result of 210° rotation

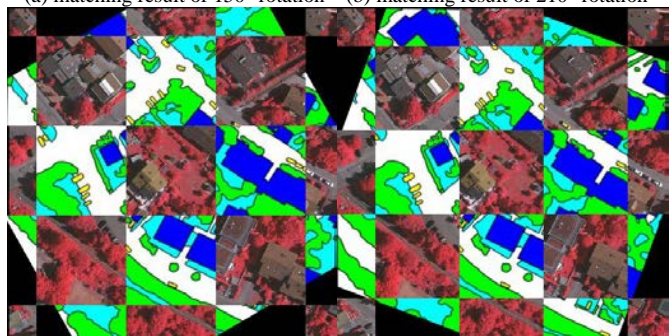

(a) registration result of 150° rotation   (b) registration result of 210° rotation

Fig. 11. Matching and registration results.

### D. Matching Performance Test

**Qualitative comparisons:** we select the first image pairs from the six multi-modal datasets for evaluation, as shown in Fig. 9. Among them, Fig. 9(a) contains translation, small rotation, and small scale changes; Fig. 9(b) suffers from a translation change and a 90° rotation change; Fig. 9(c), Fig. 9(e), and Fig. 9(f) suffer from both translation and rotation changes; Fig. 9(d) only contains a translation change. Since these image pairs are all multi-modal image pairs, the imaging mechanism of them is quite different, and these image pairs contain severe NRD. Therefore, matching on these image pairs is very challenging. Fig. 12~Fig. 14 plot the results of SIFT, SAR-SIFT and the proposed RIFT, respectively.

As can be seen, SIFT algorithm fails to match on the image pairs in Fig. 12(a), Fig. 12(b), and Fig. 12(d), and matches successfully on the image pairs of Fig. 12(c), Fig. 12(e), and Fig. 12(f). The SR accuracy is 50%. However, even if the matching is successful, the NCMs are also small, i.e., 26, 27, and 15, respectively. Because SIFT algorithm uses gradient histograms for feature description, the matching results depend heavily on the similarity of the gradient maps of the image pair. The above analysis shows that the gradient map is very sensitive to NRD, which is the fundamental reason for its poor matching performance on multi-modal images. SAR-SIFT algorithm fails to match on the image pairs in Fig. 13(a), Fig. 13(b), Fig. 13(d), and Fig. 13(e), and matches successfully on the image pairs of Fig. 13(c) and Fig. 13(f). The SR accuracy is only 33.3%. Similarly, the NCMs for SAR-SIFT are also small, which are 8 and 25, respectively. Although SAR-SIFT redefines the concept of gradient to fit the SAR image matching task, the redefined gradient is even more sensitive to NRD. In addition, SAR-SIFT uses a multiscale Harris detector for feature detection. The detector usually obtains fewer feature points and can not even get any keypoints on some images. For example, the number of feature points detected by SAR-SIFT on the google map of Fig. 13(e) is 0, resulting in the NCM must be 0. In contrast, the proposed RIFT algorithm matches successfully on all six image pairs, and the matching SR accuracy is 100%. The NCMs of RIFT are large, which are 47, 295, 324, 94, 80 and 92, respectively. The average NCM for RIFT is about 6.7 times that of SIFT and 9 times that of SAR-SIFT.

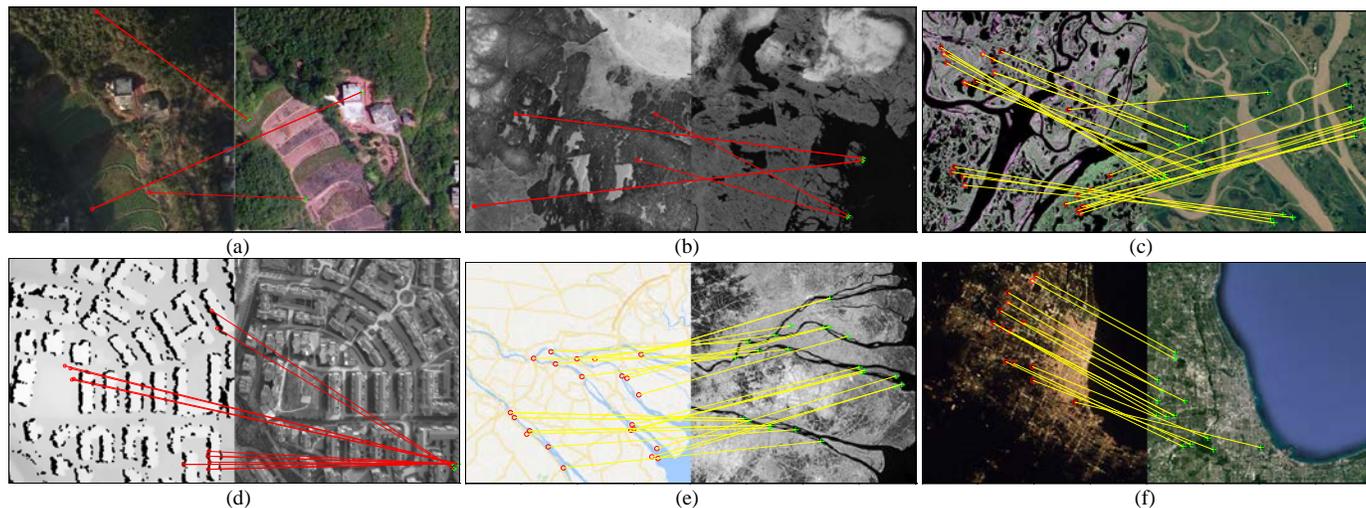

Fig. 12. Feature matching results of SIFT. The red circles and the green crosshairs in the figure indicate the feature points on the reference image and the target image, respectively; the yellow lines and the red lines indicate the correct matches and the outliers, respectively.

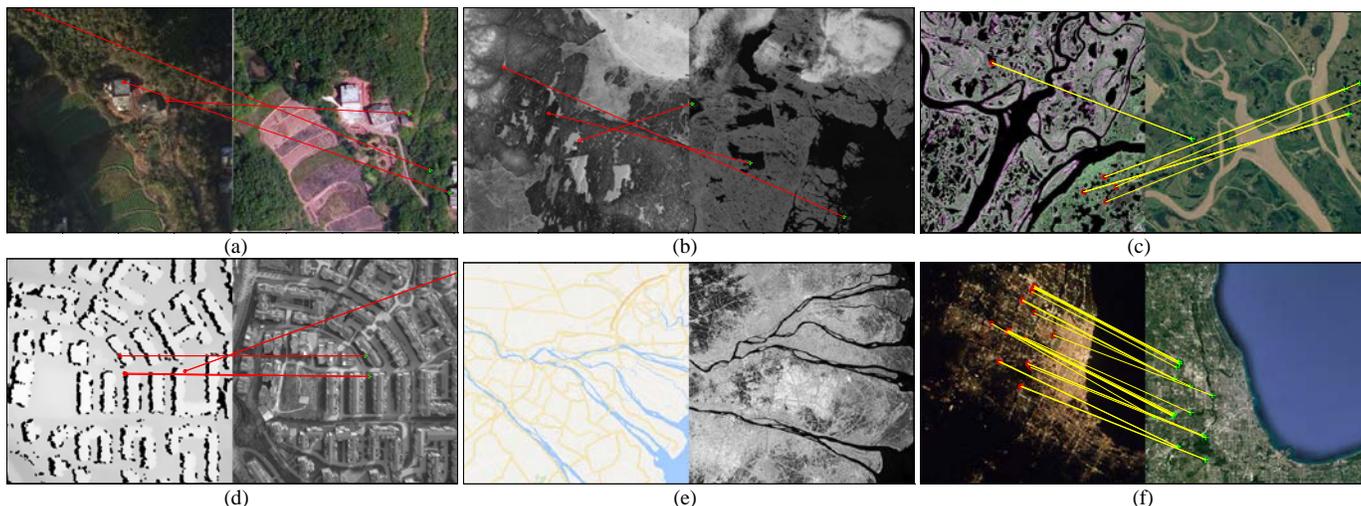

Fig. 13. Feature matching results of SAR-SIFT. The red circles and the green crosshairs in the figure indicate the feature points on the reference image and the target image, respectively; the yellow lines and the red lines indicate the correct matches and the outliers, respectively.

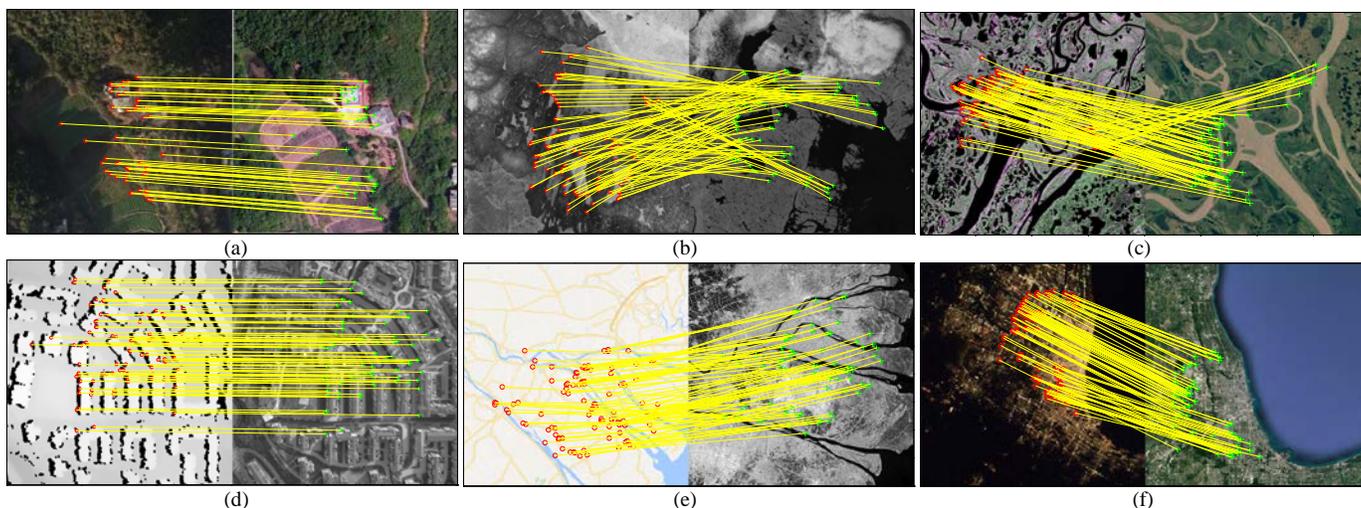

Fig. 14. Feature matching results of the proposed RIFT. The red circles and the green crosshairs in the figure indicate the feature points on the reference image and the target image, respectively; the yellow lines and the red lines indicate the correct matches and the outliers, respectively.

The matching performance of RIFT on the image pairs with NRD is far superior to the current popular feature matching methods. There are two main reasons: (1) RIFT uses PC map instead of image intensity for feature detection, and considers both the feature repetition rate and the feature number, which lays a foundation for subsequent matching. (2) RIFT adopts the log-Gabor convolution sequence to construct MIM instead of gradient map for feature description. MIM has very good robustness to NRD, thus ensuring the accuracy of the feature vectors. Fig. 17 shows more experimental results of the proposed algorithm.

**Quantitative comparisons:** Fig. 15 is the quantitative results of NCM metric, where Fig. 15(a)~Fig. 15(f) show the results of the three comparison methods on the six multi-modal datasets, respectively. As can be seen, SIFT performs better on the optical-optical dataset and the day-night dataset than the other 4 datasets because of its resistance to illumination changes. In the optical-optical dataset, the difference in imaging mechanism between images is smaller than that of the other four data sets, and matching is relatively easy. The day-night dataset is also essentially an optical-optical dataset.

The difference is the light conditions of day-night dataset are more complex. SIFT performs the worst on the depth-optical dataset. The SR accuracy is zero, and no correct matches are obtained. The reasons may be as follows: (1) SIFT uses gradient information for feature description. Gradient can reflect the structural information (edge information) of an image to a certain extent. However, in depth map or disparity map, the edge structure is relatively weak. (2) SIFT detects feature points directly based on intensity. The number of extracted feature points is small and the distribution is poor, (especially in depth maps and disparity maps, as shown in Fig. 1), resulting in poor matching performance. In most of successful matched image pairs, the NCMs of SIFT are very small (smaller than 50). In some images, there are even only a few correct matching points. The performance of SAR-SIFT is similar to that of SIFT algorithm, and its performance on the optical-optical dataset and the SAR-optical dataset is superior to the other 4 data sets. As described above, the difference in the imaging mechanism between the image pairs of optical-optical dataset is relatively small. Because SAR-SIFT algorithm is specifically designed for SAR image matching and the gradient concept is redefined, it may be more suitable for SAR-optical dataset. SAR-SIFT performs the worst on the infrared-optical dataset and the map-optical dataset, and fails completely. The radiation characteristics of the infrared-optical datasets are quite different, and the radiation characteristics of most objects are completely opposite. As shown in Fig. 9(b), black objects in the optical image appear white in the infrared image. Therefore, the redefined gradient may be more sensitive to this inverse difference. As previously analyzed, multi-scale harris detector is difficult to extract feature points on the map of map-optical dataset, which will inevitably lead to matching failure. The NCMs of SAR-SIFT are also very small in most of successful matched image pairs. However, on a few image pairs, such as image pair 9 of the SAR-optical dataset and image pair 8 of the day-night dataset, the NCMs obtained by SAR-SIFT are even larger than the proposed RIFT. In general, the matching performance of SAR-SIFT is extremely unstable. In contrast, the proposed RIFT successfully matches all the image pairs of the six datasets, and the NCMs are much greater than 50 on most of the image pairs. The matching performance of RIFT is very stable and robust, and it is hardly affected by the type of radiation distortions. RIFT is far superior to SIFT and SAR-SIFT.

Table 5 summarizes the matching SRs of the three comparison methods on each data set. As shown, SIFT has the highest SR on the day-night dataset, which is 60%; the SRs of SAR-SIFT on the optical-optical dataset and the SAR-optical dataset are both 50%; and the SRs of the proposed RIFT on all datasets are 100%. The average SRs of SIFT, SAR-SIFT, and RIFT on all six datasets are 31.7%, 28.3%, and 100%, respectively. Compared with SIFT and SAR-SIFT, the proposed RIFT improves by 68.3 and 71.7 percentages, respectively. Fig. 16 plots the average ME and RMSE of the proposed RIFT on each image pair. Because SIFT and SAR-SIFT have too low SR accuracy, their corresponding ME and RMSE are not calculated. As can be seen, the ME of RIFT is between 1.2 pixels and 2.1 pixels, and the RMSE is between 1.4 pixels and 2.2 pixels. There are many reasons for these errors, such as ground truth geometric model estimation error, estimated geometric model error, and the error of feature point positioning. Table 6 reports the NCM, ME, and RMSE of the proposed RIFT on each dataset. From the table, the NCMs of RIFT are relatively large and very stable, all of which are about 100; the matching precision is high, where the ME is about 1.8 pixels, and the RMSE is about 1.9 pixels. As mentioned earlier, RIFT is almost not affected by the type of radiation distortions. The average NCM, ME, and RMSE over all 60 image pairs are 122.4, 1.79 pixels, and 1.94 pixels, respectively.

Summarizing the above qualitative and quantitative experimental results, we can draw the following conclusions: The proposed algorithm is specially designed for NRD problems, including feature detection and feature description. Therefore, the proposed algorithm has very good resistance to NRD, and is almost not affected by the type of radiation distortions. The proposed method has achieved very good NCMs and matching accuracy on all six datasets. The matching performance of RIFT is far superior to the current classical feature matching methods. The proposed RIFT is a feature matching algorithm that has rotation invariance and is suitable for a variety of multi-modal remote sensing images.

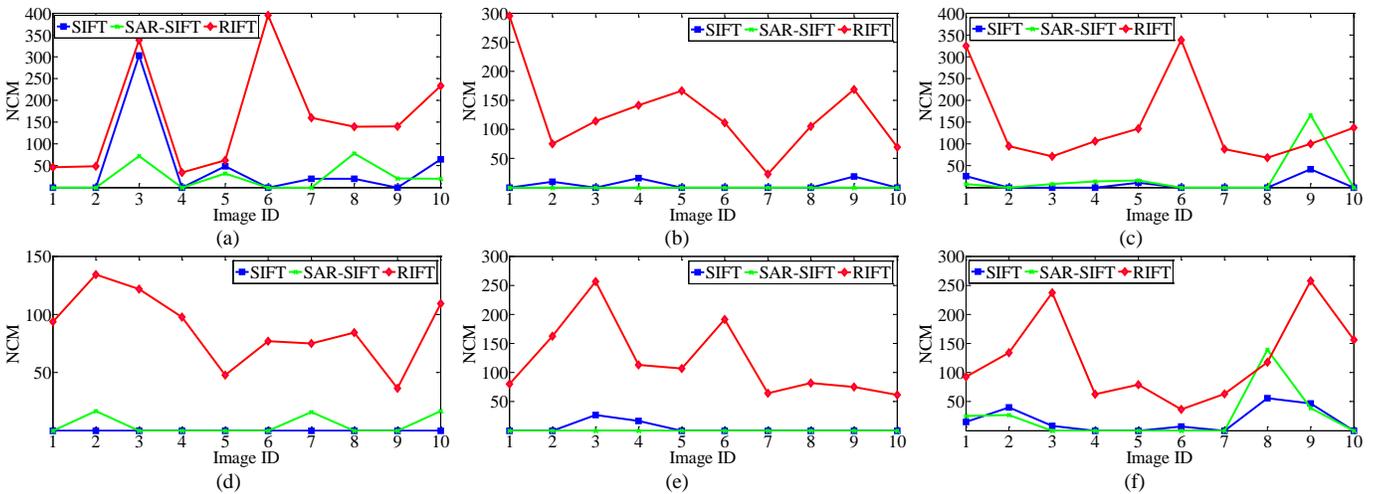

Fig. 15. Comparisons on NCM metric.

Table 5 Comparisons on SR metric.

| method | SR/% | | | | | |
|---|---|---|---|---|---|---|
| | optical-optical | infrared-optical | SAR-optical | depth-optical | map-optical | day-night |
| SIFT | 50 | 30 | 30 | 0 | 20 | 60 |
| SAR-SIFT | 50 | 0 | 50 | 30 | 0 | 40 |
| RIFT | 100 | 100 | 100 | 100 | 100 | 100 |

Table 6 Quantitative evaluation results of RIFT.

| metric | optical-optical | infrared-optical | SAR-optical | depth-optical | map-optical | day-night |
|---|---|---|---|---|---|---|
| NCM | 153.70 | 122.10 | 140.82 | 84.3 | 114.8 | 118.9 |
| ME/pixels | 1.66 | 1.77 | 1.86 | 1.83 | 1.81 | 1.83 |
| RMSE/pixels | 1.82 | 1.92 | 1.99 | 1.98 | 1.95 | 1.97 |

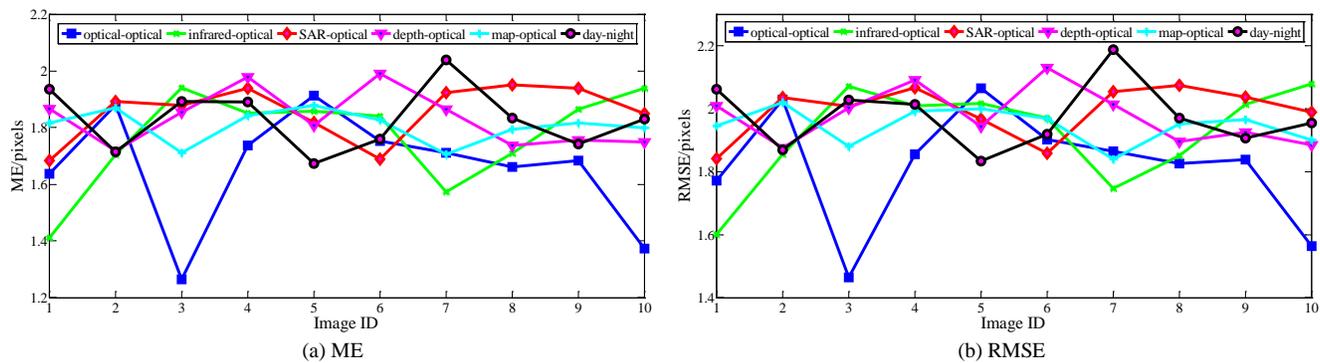

(a) ME

(b) RMSE

Fig. 16. The ME and RMSE of the proposed RIFT.

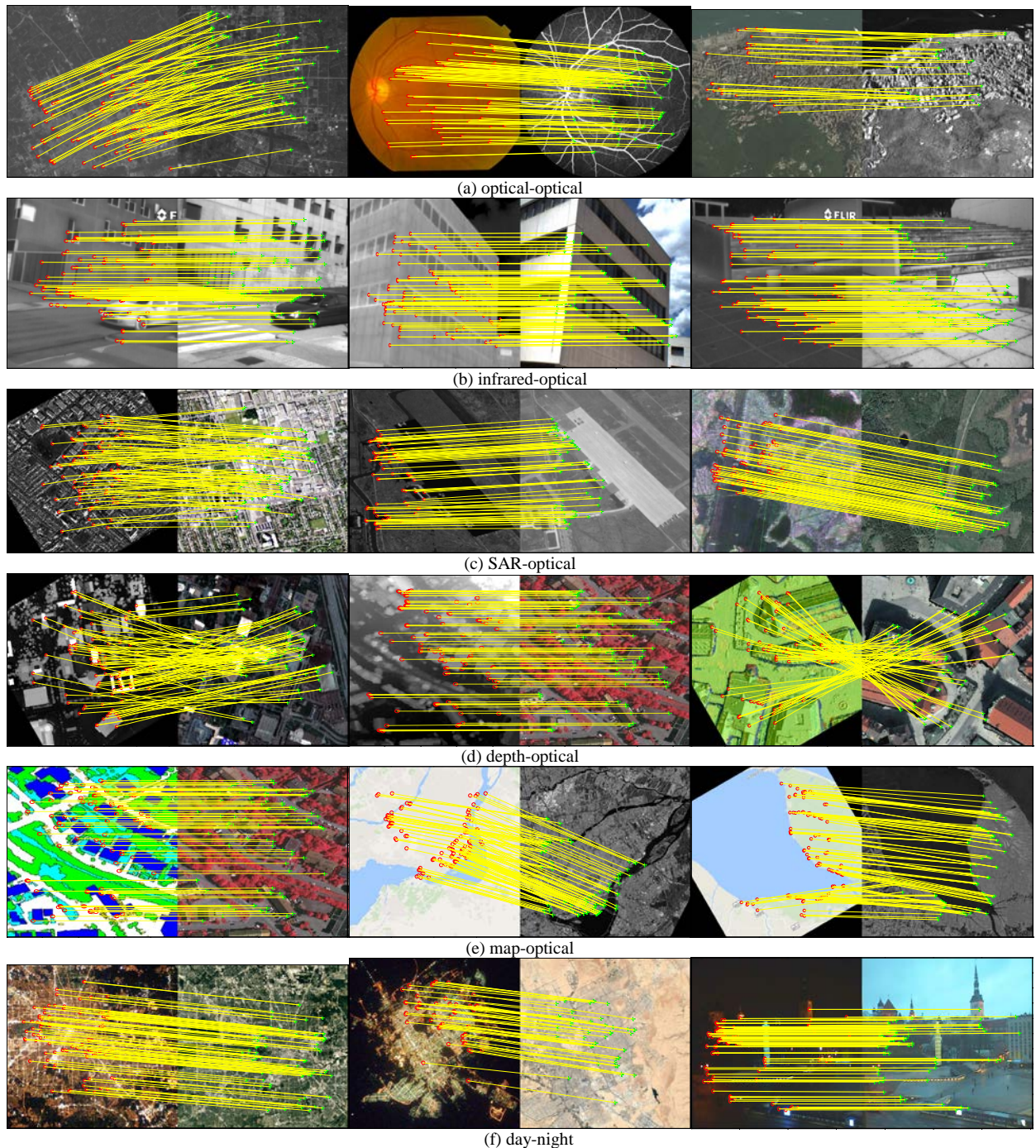

(a) optical-optical

(b) infrared-optical

(c) SAR-optical

(d) depth-optical

(e) map-optical

(f) day-night

Fig. 17. More results of the proposed RIFT.

## V. CONCLUSION

In this paper, we propose a radiation-invariant feature matching method called RIFT. This method has rotation invariance and is suitable for a variety of multi-modal remote sensing images. We first introduce the concept of PC. The initial motivation of RIFT is derived from the fact that PC has good radiation robustness. After analyzing and summarizing the drawbacks of current methods, we describe the details of RIFT. In feature detection, we obtain both corner features and edge features based on PC map, taking into account the number of features and repetition rate. We propose a MIM instead of gradient for feature description, which has very good robustness to NRD. We also analyze the inherent influence of rotations on the values of MIM, and achieve rotation-invariant by construction of multiple MIMs.

In the experiment, we first study the parameters and test the rotation invariance of RIFT; then, we make qualitative and quantitative comparisons on six datasets to verify the reliability and superiority of RIFT; finally, we analyze the limitations of RIFT and give some ideas for further improvement.